\documentclass[11pt]{article}

\usepackage[preprint]{acl}
\usepackage{amsmath,amssymb,amsfonts}
\DeclareMathOperator*{\argmax}{arg\,max}
\DeclareMathOperator*{\argmin}{arg\,min}
\usepackage{pifont}
\usepackage{tikz}
\usepackage{tcolorbox}
\tcbuselibrary{breakable}
\tcbuselibrary{listings}
\newcommand*\circled[1]{\tikz[baseline=(char.base)]{
            \node[shape=circle,draw,inner sep=0.4pt] (char) {#1};}}
\usepackage{times}
\usepackage{booktabs}
\usepackage{latexsym}
\usepackage{cancel}
\newcommand{\xmark}{\ding{55}}

\usepackage[T1]{fontenc}

\usepackage[utf8]{inputenc}

\usepackage{microtype}

\usepackage{inconsolata}

\usepackage{graphicx}

\title{Optimizing Conversational Quality in Spoken Dialogue Systems with Reinforcement Learning from AI Feedback}

\author{Siddhant Arora$^{1}$, Jinchuan Tian$^{1}$, Jiatong Shi$^{1}$, Hayato Futami$^{2}$, \\ {\bf Yosuke Kashiwagi$^{2}$, Emiru Tsunoo$^{2}$, Shinji Watanabe$^{1}$}\\
$^{1}$ Carnegie Mellon University, USA\\ 
$^{2}$ Sony Group Corporation, Japan\\
  \texttt{\{siddhana\}@cs.cmu.edu} \\
}

\begin{document}
\maketitle
\begin{abstract}
Reinforcement learning from human or AI feedback (RLHF/RLAIF) for speech-in/speech-out dialogue systems (SDS) remains underexplored, with prior work largely limited to single semantic rewards applied at the utterance level. Such setups overlook the multi-dimensional and multi-modal nature of conversational quality, which encompasses semantic coherence, audio naturalness, speaker consistency, emotion alignment, and turn-taking behavior. Moreover, they are fundamentally mismatched with duplex spoken dialogue systems that generate responses incrementally, where agents must make decisions based on partial utterances. We address these limitations with the first multi-reward RLAIF framework for SDS, combining semantic, audio-quality, and emotion-consistency rewards. To align utterance-level preferences with incremental, blockwise decoding in duplex models, we apply turn-level preference sampling and aggregate per-block log-probabilities within a single DPO objective.
We present the first systematic study of preference learning for improving SDS quality in both multi-turn Chain-of-Thought and blockwise duplex models, and release a multi-reward DPO dataset to support reproducible research. Experiments show that single-reward RLAIF selectively improves its targeted metric, while joint multi-reward training yields consistent gains across semantic quality and audio naturalness. These results highlight the importance of holistic, multi-reward alignment for practical conversational SDS.
\end{abstract}

\section{Introduction}
\begin{figure*}[t]
\centering
    \includegraphics[width=\linewidth]{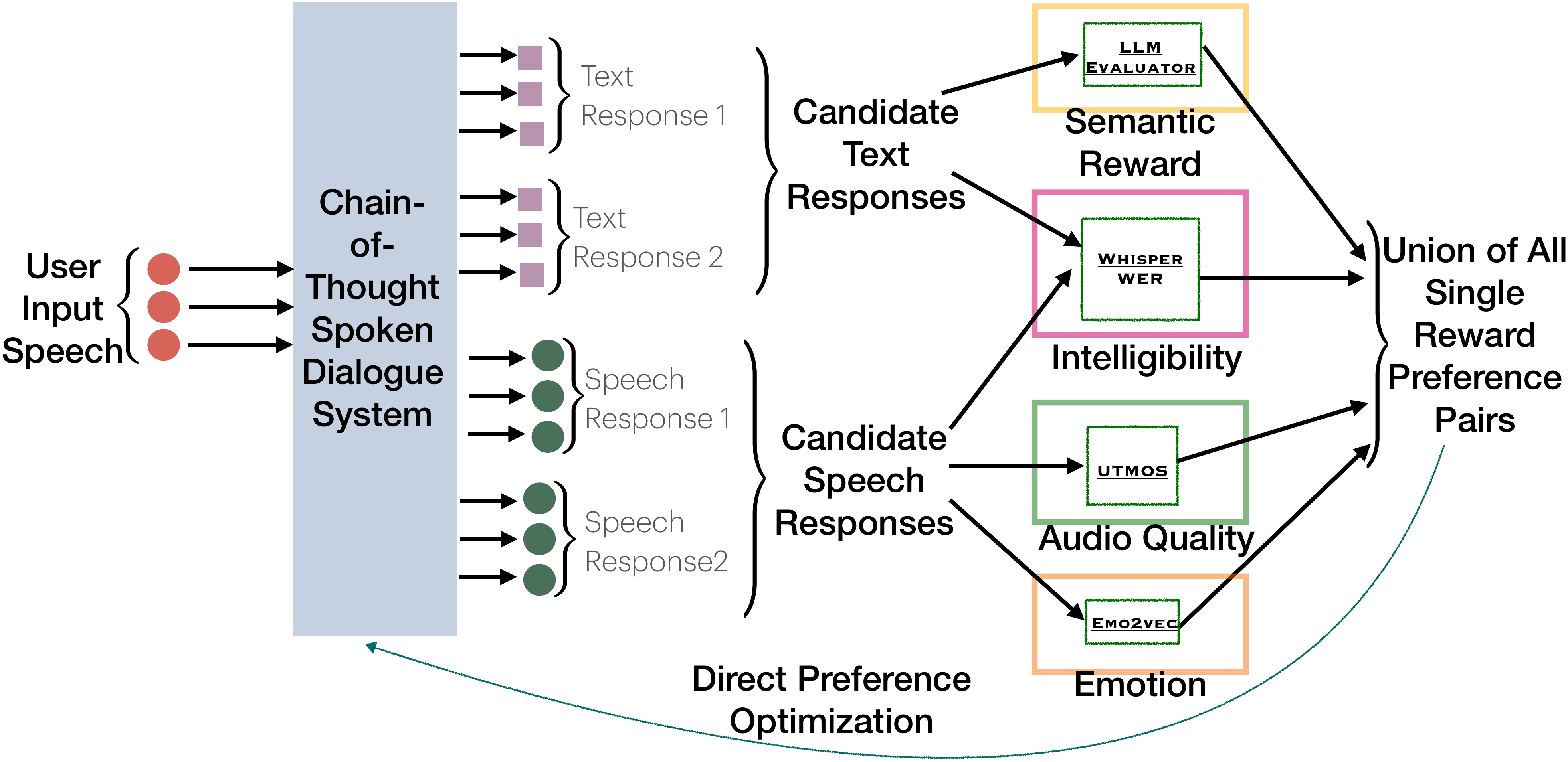} %
\caption{Overview of the proposed dataset-level Multi-Reward RLAIF framework. Unlike prior work that optimizes single rewards (typically semantic quality), we construct four independent preference datasets targeting semantic coherence, audio naturalness, intelligibility, and emotion consistency. These datasets are jointly sampled during DPO training. The framework supports both turn-by-turn CoT and blockwise duplex SDS by aggregating per-block log-probabilities under utterance-level preferences (S.~\ref{subsec:blockwise_dpo})}
\vskip -0.2in
\label{fig:proposed-cot-multiplexing-approach}
\end{figure*}
Spoken dialogue systems (SDS) are rapidly evolving from turn-based voice assistants~\cite{huggingface_speech_to_speech,xie2024miniomnilanguagemodelshear} toward real-time, full-duplex conversational agents~\cite{Dialog_GSLM,meng2024parrot,zhang2024turnbasedgameenablingrealtime,kyutai2024moshi} capable of listening and speaking simultaneously. Advances in end-to-end architectures, speech foundation models~\cite{kimiteam2025kimiaudiotechnicalreport,jinchuan2024speechlm,arora2025landscapespokenlanguagemodels}, and duplex decoding~\cite{veluri2024beyond,zhang2024omniflatten} have enabled more natural interactions, reduced latency, and improved conversational flow. Yet, despite these modeling breakthroughs, achieving human-preferred conversational behavior, semantic coherence, natural prosody, emotionally aligned delivery, and responsive turn-taking, remains an open challenge~\cite{fullduplexv2}. As systems become more sophisticated, small errors in timing, prosody, or semantic drift can accumulate across turns, directly degrading user experience.

These challenges highlight a growing alignment gap between advances in duplex modeling and the ability to optimize SDS for human-preferred conversational behavior. While there have been efforts to apply reinforcement learning from human or AI feedback (RLHF / RLAIF) as a post-training mechanism in text-based dialogue models and cascaded systems~\cite{su2016online,serban2018deep}, systematic investigations of preference learning for end-to-end (E2E) spoken dialogue systems remain scarce. To address this gap, recent work~\cite{chen2024enhancingzeroshottexttospeechsynthesis,zhang2024speechalign,cao2025mosrmb} has begun applying RLHF / RLAIF to speech-based systems, but each explores only a narrow slice of the alignment space. One such work~\cite{wu2025aligning} shows that large-scale semantic preference learning can improve factuality, safety, and coherence in a deployed duplex SDS by transcribing real user–agent dialogues and using an LLM judge to generate preference data. Align-SLM~\cite{lin-etal-2025-align} demonstrates that LLM-based semantic evaluation can effectively guide preference optimization in speech-to-speech language models, producing more coherent and less repetitive continuations. ORISE~\cite{chen2025reinforcement} focuses on temporal alignment, using online RL with heuristic, audio-driven rewards to improve turn-taking behavior and responsiveness without requiring labeled data. Together, these works illustrate the promise of RLHF for SDS.

However, these approaches still fall short of what real-time SDS demand. First, all existing methods optimize a single reward dimension, typically semantic quality or turn-taking, even though conversational quality is inherently multi-objective, jointly shaped by semantics, audio naturalness, speaker style consistency, emotional tone, and responsiveness. Second, most current RLHF pipelines assume full-utterance feedback, which is incompatible with duplex SDS where the model must commit to partial utterances and make in-flight decisions before sentence boundaries are detected. Third, while audio-based preference learning has recently gained traction in text-to-speech (TTS) systems~\cite{chen2024enhancingzeroshottexttospeechsynthesis,tian2025preference}, it remains largely unexplored in the context of SDS. Existing SDS-focused RLHF approaches primarily optimize semantic or timing-related objectives and do not incorporate audio-level rewards, such as MOS-based quality estimators or emotion similarity with human reference responses, within a unified dialogue-level learning framework. As a result, the potential of audio-centric rewards to jointly improve conversational quality in SDS has yet to be systematically studied.

To overcome these limitations, \circled{1} we propose the first multi-reward RLAIF framework for speech-in/speech-out SDS, designed to jointly optimize semantic quality, audio naturalness (UTMOS) and intelligibility and emotion fidelity within a single preference-learning pipeline. 
\circled{2} We address the mismatch between incremental decoding in duplex SDS and sentence-level rewards by introducing a streaming preference learning formulation that applies utterance-level feedback to partial, blockwise generation, enabling RLAIF to shape in-flight decisions in duplex streaming models~\cite{SCoT}.
Applied to multi-turn Chain-of-Thought SDS models and duplex architectures, our framework delivers consistent improvements across semantic and acoustic metrics. This work establishes a unified foundation for holistic alignment of SDS, moving beyond semantic-only optimization toward agents that sound better and interact more naturally. \circled{3} Finally, we will publicly release the first large-scale multi-reward preference learning dataset for SDS, together with all training and inference code, enabling reproducible RLAIF research for speech-in/speech-out conversational agents.

\section{Problem Formulation}
As modelled in prior works~\cite{SCoT}, SDS take a $d$-dimensional continuous feature stream $X = (\mathbf{x}_t \in \mathbb{R}^d \mid t = 1, \dots, T)$ as input audio from the user and generate a synchronized spoken response $Y^{\text{sds}} = (\mathbf{y}^{\text{sds}}_t \in \mathbb{R}^d \mid t = 1, \dots, T)$, where $T$ is the total duration of the conversation, including multiple turns. The objective of an SDS is to generate speech outputs that are coherent, natural, and contextually appropriate, typically modeled by estimating the conditional distribution 
$P(Y^{\text{sds}}|X, X^{\text{spk}})$, where $X^{\text{spk}}$ represents a speaker or style prompt controlling vocal characteristics (e.g., timbre, prosody, or emotion)

\section{Spoken dialogue systems}
\subsection{Turn by turn SDS}
\label{subsec:turn-by-turn}
Traditional SDS~\cite{glass1999challenges} are commonly designed around a turn-based interaction paradigm, in which the system alternates between listening and speaking. 
Under this formulation, the continuous input–output streams are partitioned into a sequence of turns, i.e. $Y^{\text{sds}}=\{Y^{\text{sds}}_k \mid k=1, \dots, K\}$ and $X=\{X_k \mid k=1, \dots, K\}$ where $K$ denotes the total number of turns.
Given this turn-based interaction, the system models the overall speech output distribution using the causality-based conditional independence (C.I.) assumption:
\begin{equation}
    P(Y^{\text{sds}}|X, X^{\text{spk}})= \prod_{k} P(Y^{\text{sds}}_k|Y^{\text{sds}}_{1:k-1},X_{1:k},X^{\text{spk}}) \label{eq:intro_sds_turn_CI}
\end{equation}
Thus, the objective of a turn-by-turn SDS is to predict, at each turn, the response that maximizes this turn-level conditional probability.

Prior work~\cite{arora2025cotsds} has explored incorporating structured intermediate representations, such as ASR transcripts $S^\text{asr}_{k}$ and textual responses $S^\text{res}_{k}$, within an E2E SDS framework, yielding Chain-of-Thought (CoT) E2E SDS models. Further details on the mathematical formulation of the CoT SDS proposed in ~\cite{arora2025cotsds} are in S.~\ref{subsec:appendix_cot_t2t_sds}. We adopt this CoT formulation as the turn-by-turn SDS backbone in our experiments.

\subsection{Duplex SDS}
\label{subsec:duplex_SDS}
A growing body of work has explored duplex, or simultaneous speaking-while-listening, spoken dialogue systems, motivated by the need to support more natural, low-latency interactions that better resemble human conversational behavior~\cite{EOT1,turntakeingIS18,skantze-2017-towards,MEENA2014903,EOT2}. 

One common strategy for building duplex SDS is the time-multiplexing approach, which alternates between processing \emph{fixed}-duration segments of user input, and generating spoken output for the subsequent segment~\cite{veluri2024beyond}.
Building on earlier turn-by-turn CoT formulations for E2E SDS, recent work, namely SCoT, has investigated how structured intermediate reasoning can be integrated into blockwise duplex systems~\cite{SCoT}. Further details on the mathematical formulation of SCoT~\cite{SCoT} can be found in S.~\ref{subsec:appendix_cot_duplex_sds}. In this work, we build on SCoT-style duplex models and focus on aligning blockwise generation with utterance-level preference learning.

\section{RLAIF Post-training Framework}
Having introduced both turn-based and duplex SDS architectures, we now describe how these models are aligned with human-preferred conversational behavior through preference optimization. The specific reward functions employed in this work are detailed later in Section~\ref{sec:multi_reward}.
\subsection{Preference Optimization for Turn-by-Turn SDS}
\label{subsec:dpo_turn}

While the CoT-based SDS formulation~\cite{arora2025cotsds} provides a structured autoregressive model for predicting turn-level responses, it does not guarantee that the generated speech aligns with human conversational preferences. To address this limitation, we adopt reinforcement learning from AI feedback (RLAIF) paradigm~\cite{bai2022constitutional,lee2023rlaif} to simulate preference signals, and apply Direct Preference Optimization (DPO)~\cite{rafailov2023direct} as a post-training objective. This enables alignment of the turn-level SDS policy with multi-reward preference signals spanning semantic, acoustic, and stylistic dimensions (see Section~\ref{sec:multi_reward}).

\paragraph{Turn-Level Preference Pairs.}
For each user turn $k$, we generate a set of $n$ candidate spoken responses
\[
\mathcal{Y}_k = \{Y^{\text{sds}}_{k,1}, \dots, Y^{\text{sds}}_{k,n}\}.
\]
Using multi-reward scoring, we select one or more preferred--dispreferred response pairs from this candidate set, forming preference pairs
\[
(Y_k^{+}, Y_k^{-}) \in \mathcal{Y}_k \times \mathcal{Y}_k,\qquad Y_k^{+} \succ Y_k^{-},
\]
where $Y_k^{+}$ and $Y_k^{-}$ denote the positive and negative responses according to the reward criteria.

\paragraph{DPO Loss for Turn-Level SDS.}
Given a preference pair $(Y^+_k, Y^-_k)$ at turn $k$ with turn history $\mathcal{H}^{(k)} = \big( X_{1:k},\, Y^{\text{sds}}_{1:k-1}, X^{\text{spk}} \big)$, the Direct Preference Optimization (DPO) loss is defined as
\begin{align}
\mathcal{L}_{\text{DPO}}^{(k)}
&=
- \log \sigma\!\Big(
\beta \big[
\Delta_\theta(k) - \Delta_{\text{ref}}(k)
\big]
\Big),
\label{eq:dpo_main_methods}
\end{align}
where
\begin{multline}
\Delta_\theta(k)
=
\log \pi_\theta\!\left(
Y^+_k \mid \mathcal{H}^{(k)}
\right)
- \\
\log \pi_\theta\!\left(
Y^-_k \mid \mathcal{H}^{(k)}
\right),\label{eq:dpo_delta}
\end{multline}
and $\Delta_{\text{ref}}(k)$ is defined analogously using the frozen reference policy $\pi_{\text{ref}}$.
Here, $\pi_\theta$ denotes the learnable SDS policy being optimized, while $\pi_{\text{ref}}$ is a fixed copy of the same model \emph{before} preference learning (i.e., the supervised fine-tuned SDS checkpoint). The reference policy serves as an anchor that stabilizes optimization by preventing large deviations from the original behavior, following standard DPO practice. The function $\sigma(\cdot)$ denotes the logistic sigmoid, and $\beta$ controls the sharpness of the preference margin.

\subsection{Preference Optimization for Duplex SDS}
\label{subsec:blockwise_dpo}

Although preference labels are assigned only at the \emph{utterance level},  duplex SDS models like SCoT generate partial speech segments in a \emph{blockwise} manner.  
To ensure that utterance-level DPO training remains compatible with blockwise generation, we decompose each candidate spoken response into fixed-size blocks and compute its log-probability by aggregating blockwise contributions.

\paragraph{Blockwise Representation.}
Given a turn-level response $Y^{\text{sds}}_k$, we partition it into blocks of size $N_{\text{block}}$, i.e. $Y^{\text{sds}}_k = \{Y^{\text{sds}}_{k,b} \mid b = 1, \dots\}$.
This mirrors the block sequence used in duplex SDS for the user speech $X$ as discussed in S.~\ref{subsec:appendix_cot_duplex_sds}.

\paragraph{Blockwise Factorization of the Utterance Probability.}
To simplify notation, we define the \emph{turn history} up to block $b$ as
\[
\mathcal{H}^{(k)}_{b}
=
\mathcal{H}^{(k)}\;\cup\; \big(
X_{k,1:b-1}, Y^{\text{sds}}_{k,1:b-1}
\big)
\]
which augments the turn history $\mathcal{H}^{(k)}$ with all user input and system output blocks generated prior to block $b$ within the current turn.
Under the same causality-based conditional independence assumption used in time-multiplexed SDS such as SCoT~\cite{SCoT},  
the posterior over the full response factorizes block-by-block as:
\begin{multline}
\log \pi_\theta(Y^{\text{sds}}_k 
\mid X_{1:k}, Y^{\text{sds}}_{1:k-1})
=\\
\sum_{b}
\log \pi_\theta\!\left(
Y^{\text{sds}}_{k,b}
\mid 
\mathcal{H}^{(k)}_{b}
\right),
\label{eq:blockwise_logprob_sum}
\end{multline}
This formulation enables us to use preference pairs $(Y_k^{+}, Y_k^{-})$ are constructed at the utterance level, as in turn-by-turn SDS (S.~\ref{subsec:dpo_turn}), and no reward labels are assigned to individual blocks. During training, the utterance-level preference is applied by substituting the aggregated log-probabilities in Eq.~\ref{eq:blockwise_logprob_sum} into the DPO objective (Eq.~\ref{eq:dpo_delta}). 
This strategy avoids the need to define reward signals at the partial-utterance level, an aspect that is non-trivial for most conversational and perceptual metrics, while still ensuring that every block contributes to optimizing the preferred response.

\section{Reward Functions}
\label{sec:multi_reward}
In this work, we adopt a dataset-level formulation of multi-reward DPO. Rather than defining a single preference pair using multiple reward constraints simultaneously, we construct independent preference datasets, each targeting a specific dimension of conversational quality (semantic coherence, audio quality, intelligibility, emotion consistency). These datasets are then jointly used during DPO post-training, enabling the SDS policy to be optimized across complementary objectives within a unified framework.
\subsection{Semantic Quality Reward}
\label{subsec:sem_multi_reward}
To assess the semantic quality of generated spoken responses, we adopt the LLM-based semantic evaluation framework introduced in Align-SLM~\cite{lin-etal-2025-align} and adapt it to the SDS setting. For each user turn, the CoT SDS samples multiple candidate intermediate text responses $\hat{S}^{\text{res}}_k$ (Eq.~\eqref{eq:turn_cot_t2t_formulation} in Appendix~\ref{subsec:appendix_cot_t2t_sds} for more details), which are scored by Qwen2.5-72B-Instruct~\cite{qwen2025qwen25technicalreport} as the LLM judge to construct DPO data (See~\ref{box:llm_judge_prompt} for LLM prompt). 
The judge rates each candidate on coherence, relevance and grounding with dialogue context, producing a scalar semantic score between 0 and 10. In parallel, we follow prior work~\cite{lin-etal-2025-align} and also compute an \emph{AutoBLEU}~\cite{lakhotia2021generative} score for each candidate, which serves as an automated filter: high AutoBLEU values indicate repetitive phrasing or low-diversity text that lacks meaningful semantic content. Candidates with an LLM coherence score above a positive threshold $\tau_{\text{pos}}$ \emph{and} AutoBLEU below a repetition threshold $\delta_{\text{low}}$ are labeled as \emph{positive} samples. Conversely, candidates with LLM scores below a negative threshold $\tau_{\text{neg}}$ or AutoBLEU exceeding $\delta_{\text{high}}$ are labeled as \emph{negative} samples. This procedure yields high-quality semantic preference pairs for DPO training.

The semantic judge operates on \emph{text} responses and therefore produces a preference pair
$\hat{S}^{\text{res},+}_k \succ \hat{S}^{\text{res},-}_k$ at each turn.
To apply DPO (Eq.~\eqref{eq:dpo_main_methods}) using this signal, we relate the turn-level speech policy
$\pi_\theta(Y^{\text{sds}}_k \mid X_{1:k}, Y^{\text{sds}}_{1:k-1})$ to the CoT factorization (S.~\ref{subsec:appendix_cot_t2t_sds}) in which the text response
$\hat{S}^{\text{res}}_k$ is an explicit intermediate variable.

Using the chain rule and Viterbi-style approximation, we can write the likelihood of a turn-level spoken response as
\begin{align}
\log \pi_\theta(Y_k \mid \mathcal{H}^{(k)})
\approx 
\log P_\theta(\hat{S}^{\text{res}}_k \mid \mathcal{H}^{(k)})
+ \nonumber \\
\log P_\theta(Y_k \mid \mathcal{H}^{(k)}, \hat{S}^{\text{res}}_k).
\label{eq:policy_viterbi_text}
\end{align}

Since the semantic preference signal depends only on the judged text response,
we modify Eq.~\ref{eq:dpo_delta} and apply DPO using only the \emph{text-policy term}:
\begin{multline}
\Delta_\theta^{\text{sem}}(k)
=
\log P_\theta(\hat{S}^{\text{res},+}_k \mid \mathcal{H}^{(k)}) -
\\
\log P_\theta(\hat{S}^{\text{res},-}_k \mid \mathcal{H}^{(k)}),
\label{eq:delta_text_policy}
\end{multline}
(and analogously for $\pi_{\text{ref}}$), which is then substituted into Eq.~\eqref{eq:dpo_main_methods}.
Intuitively, this treats semantic DPO as directly increasing the likelihood of the preferred
\emph{text response}\footnote{We tried other ablations, including (i) applying DPO jointly to text and speech responses and (ii) augmenting DPO with an additional supervised fine-tuning (SFT) loss on speech outputs. Across these settings, we found that a simple text-only DPO formulation consistently achieved the strongest and most stable improvements, while more complex variants did not provide additional gains.}.
\subsection{Audio Quality and Intelligibility Rewards}
\label{sec:audio_rewards}

Beyond semantic alignment, high-quality spoken dialogue systems must also produce responses that are acoustically natural and intelligible. To this end, we construct preference learning signals targeting \emph{audio quality} and \emph{intelligibility}, while carefully controlling for semantic content.

\paragraph{Audio Quality.}
To optimize acoustic naturalness, we use an automatic speech quality estimator $q(\cdot)$ (UTMOS~\cite{saeki22c_interspeech}) as a proxy for perceived audio quality.
For each dialogue turn $k$, we generate a set of candidate spoken responses
$\mathcal{Y}_k$ as in S.~\ref{subsec:dpo_turn}.
Let $q(Y^{\text{sds}}_{k,i})$ denote the predicted quality score for a synthesized utterance $Y^{\text{sds}}_{k,i}$.
We construct preference pairs by selecting
\begin{equation}
Y_k^{+} = \argmax_{Y^{\text{sds}}_{k,i} \in \mathcal{Y}_k} q(Y^{\text{sds}}_{k,i}), 
Y_k^{-} = \argmin_{Y^{\text{sds}}_{k,i} \in \mathcal{Y}_k} q(Y^{\text{sds}}_{k,i}),
\end{equation}

\paragraph{Intelligibility.}
Let $w(Y^{\text{sds}}_{k,i})$ denote the WER between the hypothesis of synthesized speech $Y^{\text{sds}}_{k,i}$ generated by pre-trained ASR system and the model-predicted text response $S^{\text{res}}_k$.
Among candidates satisfying $w(Y^{\text{sds}}_{k,i}) \le \tau_{\text{wer}}$, we select
\begin{equation}
Y_k^{+} = \argmin_{Y^{\text{sds}}_{k,i} \in \mathcal{Y}_k} w(Y^{\text{sds}}_{k,i}).
\end{equation}
and draw negative samples from candidates with
\begin{equation}
Y_k^{-} \sim \{ Y^{\text{sds}}_{k,i} \in \mathcal{Y}_k \mid w(Y^{\text{sds}}_{k,i}) \ge w(Y_k^{+}) + \delta_{\text{wer}} \}.
\end{equation}
where $\delta_{\text{wer}}$ is a hyperparameter. This margin-based construction avoids ambiguous comparisons and focuses on clearly degraded intelligibility.

For both audio quality and intelligibility rewards, we enforce that the text response $S^{\text{res}}_k$ is identical for the positive and negative samples within each DPO pair. 
Accordingly, the DPO objective for audio-based rewards can be written as
\begin{multline}
\Delta^{\text{audio}}_\theta(k)
=
\log \pi_\theta\!\left(
Y_k^{+} \mid X_{1:k}, Y^{\text{sds}}_{1:k-1}, S^{\text{res}}_k
\right)
- \\
\log \pi_\theta\!\left(
Y_k^{-} \mid X_{1:k}, Y^{\text{sds}}_{1:k-1}, S^{\text{res}}_k
\right),
\label{eq:audio_dpo_delta}    
\end{multline}
As a result, the preference signal isolates the quality of the \emph{generated speech output} alone, allowing DPO to directly improve acoustic naturalness and intelligibility without altering the underlying semantic content.

\subsection{Emotion-Consistency Reward}
\label{subsec:emotion_rewards}
In addition to semantic quality and acoustic naturalness, effective spoken dialogue systems should convey emotions consistent with human intent. To encourage emotionally aligned responses, we construct preference pairs based on \emph{emotion similarity} between synthesized speech and human reference speech. Emotion representations are extracted using a pretrained encoder, and a scalar similarity score $e(Y^{\text{sds}}_{k,i})$ is computed for each candidate.
We then select preference pairs as
\begin{equation}
Y^+_k = \argmax_{Y^{\text{sds}}_{k,i} \in \mathcal{Y}_k} e(Y^{\text{sds}}_{k,i}), 
Y^-_k = \argmin_{Y^{\text{sds}}_{k,i} \in \mathcal{Y}_k} e(Y^{\text{sds}}_{k,i}),
\end{equation}
subject to the constraint that the difference between the maximum and minimum emotion similarity exceeds a threshold $\delta_{\text{emo}}$, ensuring that the preference signal reflects a meaningful emotional contrast.
As with audio quality and intelligibility rewards, we enforce that the underlying text response $S^{\text{res}}_k$ is identical for both $Y^+_k$ and $Y^-_k$.

\begin{table}[t]
\centering
\caption{Scale of the constructed preference learning dataset used for multi-reward DPO training.}
\vskip -0.1in
\begin{tabular}{l c}
\toprule
\textbf{Reward Type} & \textbf{\# DPO Pairs} \\
\midrule
Semantic Quality  & 51.1K \\
Audio Quality & 32.0K \\
Intelligibility & 61.0K \\
Emotion & 21.6K \\
\midrule
\textbf{Total} & \textbf{165.7K} \\
\bottomrule
\end{tabular}
\vskip -0.1in
\label{tab:dpo_data_stats}
\end{table}
\begin{table*}[t]
    \centering
    \caption{Semantic Quality Evaluation of RLAIF-Post-Trained Turn-by-Turn SDS. * Statistical significant difference based on Wilcoxon signed rank and Paired Bootstrap Resampling tests with p-value<0.01. Values in bracket indicate the percentage of LLM judge responses receiving low scores (< 5). Win rate is defined only for RLAIF models and measures the proportion of samples that outperform the corresponding no-RLAIF base model.}
    \vskip -0.1in
    \resizebox{\linewidth}{!}{
    \begin{tabular}{l|cccccc}
        \hline
        \textbf{Model} & \textbf{ROUGE-L (↑)} & \textbf{Perplexity (↓)} & \textbf{AutoBLEU (↓)} & \textbf{LLM judge (↑)} & \textbf{Win Rate (↑)} \\
        \hline
        Direct E2E~\cite{arora2025cotsds} & \hphantom{0}8.4 & 302.2 & 51.5 & 5.50\hphantom{*} (24.2\%)& \xmark\\
        Moshi~\cite{kyutai2024moshi} & \hphantom{0}8.1 &  136.5 & 57.8 & 5.71\hphantom{*} (21.0\%)  & \xmark\\
        Multi turn CoT E2E~\cite{SCoT} & 12.1 & \hphantom{0}{21.2} & 68.3 & 6.18\hphantom{*} (10.2\%) & \xmark \\ \midrule
        \hphantom{00} + RLAIF (Single-Reward) & 11.9 & \hphantom{0}19.9 & \textbf{56.5} & \textbf{6.33}* (\hphantom{0}7.1\%) & 55.4  \\
        \hphantom{00} + RLAIF (Joint-Reward-v1) & 11.8 & \hphantom{0}\textbf{19.6} & 61.3 & 6.29* (\hphantom{0}8.5\%)  & 52.6 \\
        \hphantom{00} + RLAIF (Joint-Reward-v2) & 11.9 & \hphantom{0}19.9 & 59.9 & 6.33* (\hphantom{0}7.5\%) & 54.4 \\
        \hline
    \end{tabular}
    }
    \vskip -0.1in
    \label{tab:text_response_results}
\end{table*}
\section{Experiments}

\subsection{DPO Data Preparation.}
To construct high-quality preference data for DPO training, we begin by applying multi-turn CoT SDS model~\cite{arora2025cotsds} to generate diverse response candidates for real conversational contexts. We use widely adopted human--human dialogue corpora, focusing on the Switchboard dataset~\cite{Switchboard}, which contains approximately 300 hours of spontaneous telephone conversations. The constructed multi-reward DPO dataset comprises 165.7K high-quality preference pairs, with detailed statistics summarized in Table~\ref{tab:dpo_data_stats}. 
For semantic preference pairs construction (see S.~\ref{subsec:sem_multi_reward}), the CoT model generates $n=10$ candidate text responses per dialogue turn using top-k sampling, ensuring sufficient lexical and semantic diversity. For acoustic and emotion rewards (see S.~\ref{sec:audio_rewards} and S.~\ref{subsec:emotion_rewards}), we generate $n=10$ spoken realizations for each candidate text response using top-k sampling. Emotion-consistency preference pairs are constructed using emotion representations extracted by Emo2Vec~\cite{ma2023emotion2vec} from both synthesized speech and the corresponding ground-truth human speech.
Additional details are provided in S.~\ref{subsec:appendix_dpo_data_prep}.

\subsection{Evaluation Data and Metrics}
We evaluate all baslines and proposed models on the Eval2000 dataset similar to \cite{SCoT,arora2025cotsds}.
Semantic quality is evaluated by transcribing synthesized speech $Y^{\text{sds}}$ using Whisper large~\cite{whisper}.  We report ROUGE~\cite{lin2004rouge} and METEOR~\cite{banerjee2005meteor} scores against human reference responses, along with perplexity~\cite{jelinek1977perplexity} computed using GPT-2~\cite{gpt-2}. To better capture conversational relevance beyond n-gram overlap, we additionally evaluate responses using Qwen2.5-7B-Instruct as an \emph{LLM judge}, following recent evaluation protocols~\cite{zhang2024omniflatten} (Prompt in S.~\ref{box:llm_judge_prompt}). We also report \emph{AutoBLEU}, as a proxy for repetition and degeneration.
For RLAIF-trained models only, we further report a \emph{win rate} relative to the no-RLAIF baseline. Given a set of evaluated samples, the win rate is defined 
\begin{equation}
\text{WinRate}
=
\frac{
\#(s > s_{\text{base}})
+
0.5 \times \#(s = s_{\text{base}})
}{
N
},
\end{equation}
where $s$ denotes the evaluation score for a given sample, $s_{\text{base}}$ is the corresponding score from the no-RLAIF baseline, and $N$ is the total number of evaluated samples. This metric reflects the proportion of responses that outperform (or tie with) the baseline under the same evaluation criterion.
We utilize the VERSA toolkit~\cite{shi2024versa}, measuring intelligibility
through Whisper hypotheses and evaluating audio quality using UTMOS. We evaluate speaking-style consistency within entire conversations using 
Emo2Vec~\cite{ma2023emotion2vec}.
We rank all SDSs based on their emotional alignment and compute the average rank (``Emotion Rank''~\cite{arora2025cotsds}) across all utterances. 

\subsection{Models}
Following prior work~\cite{arora2025cotsds}, we first report results on a single-turn E2E (``Direct E2E'') spoken dialogue system. We further benchmark our approach against Moshi~\cite{kyutai2024moshi}, a strong 7B dual-channel duplex SDS baseline\footnote{kyutai/moshiko-pytorch-bf16}. In addition, we evaluate a multi-turn Chain-of-Thought E2E model (Multi-turn CoT E2E) introduced in prior work~\cite{SCoT}, which serves as a strong reasoning-aware baseline. We then extend our framework to the duplex setting and compare against the blockwise duplex SCoT-Response model from~\cite{SCoT}. For both the multi-turn CoT and duplex SCoT-Response models, we apply RLAIF post-training using our proposed preference-learning framework.

We conduct RLAIF post-training under multiple configurations to disentangle the effects of different reward signals. In RLAIF (Single-Reward) settings, models are trained using DPO pairs derived from a single reward signal, namely semantic quality, audio quality, intelligibility or emotion consistency, with the objective of selectively improving the corresponding metric.  In contrast, we consider two RLAIF (Joint-Reward) configurations: (i) (Joint-Reward-v1) a joint semantic–audio setting that combines semantic, audio quality, and intelligibility preference data, and (ii) (Joint-Reward-v2) a joint semantic–audio–emotion setting that additionally incorporates speaking-style (emotion-consistency) preferences. This design allows for a systematic evaluation of how individual and combined preference signals contribute to overall system performance. Additional implementation and training details are provided in the Appendix~\ref{subsec:appendix_experiment_setup}.

\section{Results}

\subsection{RLAIF with Semantic Quality Reward}
Table~\ref{tab:text_response_results} reports semantic-quality evaluation for turn-by-turn SDS, including general E2E baselines and the strong duplex SDS baseline Moshi. Among non-RLHF systems, the Multi-turn CoT E2E model substantially outperforms both the standard E2E and Moshi baselines in terms of semantic metrics, establishing a strong foundation for preference-based post-training.

Building on this baseline, RLAIF (Single-Reward) post-training with semantic preference data yields a statistically significant improvement in LLM-judge scores over the Multi-turn CoT E2E model (6.18 → 6.33, p < 0.01 under both Wilcoxon signed-rank and paired bootstrap resampling tests), with a corresponding win rate of 55.4\%. Importantly, beyond average score improvements, RLAIF induces a meaningful distributional shift: the proportion of low-quality responses (LLM-judge $<$ 5) is reduced from 10.2\% to 7.1\%, corresponding to a 28.5\% relative reduction in poor responses. This behavior is consistent with prior RLAIF findings, where improvements are often driven by suppressing degenerate or incoherent generations rather than uniformly increasing mean scores.

RLAIF post-training also improves auxiliary indicators of response quality. In particular, semantic RLAIF substantially reduces repetition, as reflected by a large drop in AutoBLEU (68.3 → 56.5), and yields more coherent responses as evidenced by lower perplexity (21.2 → 19.9). When extending to multi-objective optimization, RLAIF (Joint-Reward-v1), which combines semantic, audio quality, and intelligibility preferences, maintains strong LLM-judge performance (6.29). Incorporating additional emotion-consistency preferences in RLAIF (Joint-Reward-v2) further improves LLM-judge scores (6.33) and reduces the fraction of low-scoring responses (7.5\%), demonstrating that jointly leveraging various preferences leads to consistent overall gains without degrading semantic quality.

Together, these results show that RLAIF post-training effectively improves conversational quality in end-to-end SDS, primarily by reducing low-quality and repetitive responses while preserving semantic coherence. Based on qualitative inspection of model outputs, RLAIF post-training consistently improves direct question answering, reduces topic drift, and suppresses repetitive or degenerate response patterns, aligning well with the quantitative gains observed in LLM-judge scores and AutoBLEU. At the same time, we observe occasional failure cases where RLAIF-trained models produce responses that are overly safe or generic, and thus do not sufficiently advance the conversation. We provide representative qualitative examples in T.~\ref{tab:qualitative_better} and \ref{tab:qualitative_worse} in Appendix.

\subsection{RLAIF with Acoustic Quality Reward}
Table~\ref{tab:acoustic_response_results} reports audio quality and intelligibility evaluation across general E2E SDS baselines, the strong duplex baseline Moshi, and RLAIF-trained variants. Among non-RLAIF systems, Moshi achieves the highest absolute UTMOS scores, largely attributable to its use of high-quality assistant voice prompts. In contrast, our models are conditioned on lower-fidelity Switchboard speaker prompts, resulting in lower UTMOS scores.

Building on the Multi-turn CoT E2E baseline, RLAIF (Single-Reward) post-training demonstrates targeted and effective improvements. Training with audio-quality preferences substantially improves perceived speech naturalness, increasing UTMOS from 2.16 to 3.06. Similarly, RLAIF (Single-Reward) training with intelligibility-focused preferences yields a marked reduction in word error rate, improving turn-level intelligibility from 6.1 to 3.3. Here, WER is computed between the ASR transcript of the synthesized speech, obtained using Whisper, and the model-predicted text response $S^{\text{res}}_{k}$ used during CoT decoding (S.~\ref{sec:audio_rewards}). These results confirm that audio-centric DPO signals selectively and reliably improve their intended dimensions.

When multiple preference datasets are jointly leveraged using RLAIF (Joint-Reward) training, the model achieves simultaneous gains in both audio quality and intelligibility. In particular, RLAIF (Joint-Reward-v1), which combines semantic, audio quality, and intelligibility preferences, attains a UTMOS of 2.85 while further reducing WER to 1.0, representing the best intelligibility performance among RLAIF-trained models. Incorporating additional emotion-consistency preferences in RLAIF (Joint-Reward-v2) preserves audio quality but results in a slightly higher WER, indicating a mild trade-off between expressive consistency and intelligibility.
Together, these findings validate the effectiveness of audio-quality and intelligibility preference data and highlight the benefit of unified multi-reward RLAIF for improving speech naturalness and intelligibility.
\begin{table}[t]
    \centering
    \caption{Audio quality evaluation. \xmark: WER is not reported for Direct E2E models since they do not output explicit intermediate text response.}
    \vskip -0.1in
    \resizebox{\linewidth}{!}{
    \begin{tabular}{l|cccccc}
        \hline
        \textbf{Model} & \textbf{UTMOS (↑)} & \textbf{WER (↓)} \\
        \hline
        Direct E2E~\cite{arora2025cotsds} & 2.03 & \xmark \\
        Moshi~\cite{kyutai2024moshi} & 3.34 & \xmark \\
        Multi turn CoT E2E~\cite{SCoT} & 2.16 & 6.1 \\\midrule
        \hphantom{00} + RLAIF (Single-Reward) & \textbf{3.06} & 3.3 \\
        \hphantom{00} + RLAIF (Joint-Reward-v1) & 2.85 & \textbf{1.0}\\
        \hphantom{00} + RLAIF (Joint-Reward-v2) & 2.85 & 1.7\\
        \hline
    \end{tabular}
    }
    \vskip -0.1in
    \label{tab:acoustic_response_results}
\end{table}
\begin{table}[t]
    \centering
    \caption{Speaking style consistency evaluation. }
    \vskip -0.1in
    \resizebox{\linewidth}{!}{
    \begin{tabular}{l|cccccc}
        \hline
        \textbf{Model} & \textbf{Emotion Rank (↓)} \\
        \hline
        Direct E2E~\cite{arora2025cotsds} & 2.81\\
        Moshi~\cite{kyutai2024moshi} & 4.92\\
        Multi turn CoT E2E~\cite{SCoT} & 2.29\\\midrule
        \hphantom{00} + RLAIF (Single-Reward) & 1.98\\
        \hphantom{00} + RLAIF (Joint-Reward-v2) & 3.00\\
        \hline
    \end{tabular}
    }
    \vskip -0.1in
    \label{tab:emotion_response_results}
\end{table}
\begin{table}[t]
    \centering
    \caption{Semantic quality evaluation for duplex model. }
    \vskip -0.1in
    \resizebox{\linewidth}{!}{
    \begin{tabular}{l|cccccc}
        \hline
        \textbf{Model} & \textbf{ROUGE-L (↑)} & \textbf{Perplexity (↓)}  & \textbf{LLM judge (↑)} \\
        \hline
        \emph{SCoT-Response}  & 19.8  & \hphantom{0}42.3  & 5.95 \\
        \hphantom{00} + RLAIF & \textbf{23.1} & \hphantom{0}25.0 & 6.00  \\
        \hline
    \end{tabular}
    }
    \vskip -0.15in
    \label{tab:text_response_duplexresults}
\end{table}
\subsection{RLAIF with Emotion Quality Reward}
Table~\ref{tab:emotion_response_results} reports speaking-style consistency evaluated via emotion rank, where lower values indicate closer alignment with human reference speech. Among non-RLAIF baselines, the Multi-turn CoT E2E model performs best (2.29), outperforming both E2E and Moshi. RLAIF (Single-Reward) post-training with emotion-consistency preferences further improves alignment, reducing the emotion rank to 1.98, demonstrating the effectiveness of emotion-aware preference learning.

In contrast, RLAIF (Joint-Reward-v2) yields a higher emotion rank (3.00), indicating competition between reward signals. In particular, strong semantic rewards from LLM-based judges can encourage safer or more generic responses (T.~\ref{tab:qualitative_worse}), which may reduce emotional expressiveness. This highlights an inherent trade-off in multi-objective RLAIF settings and motivates future work on better balancing various rewards in SDS.
\subsection{RLAIF for Duplex Models}
Table~\ref{tab:text_response_duplexresults} reports semantic-quality evaluation for the duplex SDS. Compared to the SCoT-Response baseline, RLAIF post-training consistently improves semantic performance, yielding higher ROUGE-L (19.8 → 23.1), lower perplexity (42.3 → 25.0), and improved LLM-judge scores (5.95 → 6.00). These results indicate that the proposed RLAIF framework generalizes beyond turn-by-turn settings and remains effective for blockwise duplex spoken dialogue models.
\section{Conclusion}
We presented the first multi-reward RLAIF framework for E2E SDS, addressing key limitations of prior work that focuses on single, utterance-level semantic rewards. Our approach jointly optimizes semantic coherence, audio quality, intelligibility, and emotion consistency on turn-by-turn Chain-of-Thought SDS. By applying utterance-level preferences over blockwise decoding, we enable preference optimization on blockwise duplex models without requiring partial-utterance reward definitions.
Experimental results demonstrate that single-reward RLAIF selectively improves its targeted dimension, validating the specificity of our constructed preference data, while joint multi-reward training yields consistent gains across semantic and acoustic metrics. Finally, we release the first large-scale multi-reward DPO dataset for spoken dialogue systems to support reproducible research.

\section{Limitations}
While our results demonstrate the effectiveness of multi-reward RLAIF for spoken dialogue systems, several limitations remain. First, preference data construction relies on automatic evaluators, which may introduce bias or noise relative to human judgments. Incorporating human-in-the-loop preferences is an important direction for future work. Second, our multi-reward formulation combines reward signals via dataset-level concatenation rather than explicitly modeling trade-offs or interactions between objectives, which may limit optimal balancing in some conversational contexts. Finally, our experiments focus on English conversational datasets; extending the framework to multilingual settings and more diverse conversational domains remains an open challenge.

\section{Ethics Impact}
We adhere to the ACL Ethics Policy. Our experiments are based on open-source datasets with no violation of privacy, and we will make all our code and models publicly available. Parts of this manuscript were edited for clarity and language using an AI-based writing assistant. The authors take full responsibility for the content.

\bibliography{custom}

\appendix

\section{Appendix}
\label{sec:appendix}
\subsection{Mathematical Formulation of Chain-of-Thought Turn by turn SDS}
\label{subsec:appendix_cot_t2t_sds}
More recent E2E approaches replace cascaded architectures~\cite{glass1999challenges,huang2024audiogpt} with unified speech–language models (SLMs) that directly generate spoken responses autoregressively~\cite{zhang2023speechgpt,zhang2024speechgpt,nguyen2024spirit}. Although these models mitigate error propagation, they often lack explicit intermediate reasoning structure, which can limit coherence and increase data requirements~\cite{arora2025cotsds,kyutai2024moshi}.
To address this, prior work~\cite{arora2025cotsds} has explored incorporating structured intermediate representations, such as ASR transcripts $S^\text{asr}_{k}$ and textual responses $S^\text{res}_{k}$, within an end-to-end framework, yielding Chain-of-Thought (CoT) SDS models.
Using the Viterbi approximation and C.I. assumption, we can modify Eq.~\eqref{eq:intro_sds_turn_CI} to get:
\begin{multline}
P(Y^{\text{sds}}|X, X^{\text{spk}}) \\
\approx \prod_{k} P(Y^{\text{sds}}_k|Y^{\text{sds}}_{1:k-1},X_{1:k},\hat{S}^{\text{res}}_{1:k}, \hat{S}^{\text{asr}}_{1:k}, X^{\text{spk}}) 
\label{eq:turn_cot_final_formulation}
\end{multline}
where
\begin{multline}
\hat{S}^{\text{asr}}_{k} = \argmax_{S^{\text{asr}}_{k}} P(S^{\text{asr}}_{k}|Y^{\text{sds}}_{1:k-1},X_{1:k}, \\
\hat{S}^{\text{res}}_{1:k-1}, \hat{S}^{\text{asr}}_{1:k-1}, \cancel{X^{\text{spk}}})
\label{eq:turn_cot_asr_formulation}
\end{multline}
and
\begin{multline}
\hat{S}^{\text{res}}_{k} = \argmax_{S^{\text{res}}_{k}} P(S^{\text{res}}_{k}|Y^{\text{sds}}_{1:k-1},X_{1:k}, \\
\hat{S}^{\text{res}}_{1:k-1}, \hat{S}^{\text{asr}}_{1:k}, \cancel{X^{\text{spk}}})
\label{eq:turn_cot_t2t_formulation}
\end{multline}
At each turn, the \textbf{CoT model} follows a structured decoding process: where it first infers a transcription $S^{\text{asr}}_{k}$, then a textual response $\hat{S}^{\text{res}}_{k}$, and finally synthesizes speech $Y^{\text{sds}}_k$ conditioned on these intermediate variables.

\subsection{Mathematical Formulation of Chain-of-Thought Duplex SDS}
\label{subsec:appendix_cot_duplex_sds}
In the time-multiplexing setup~\cite{veluri2024beyond}, the input speech X is divided into a sequence of $B$ blocks, $X = \{X_b \mid b=1, \dots, B\}$. 
Similarly, the output speech $Y^{\text{sds}}$ is represented as a sequence of $B$ blocks, $Y^{\text{sds}} = \{Y^{\text{sds}}_b \mid b=1, \dots, B\}$.
Rather than generating responses only at turn boundaries or frame-by-frame, the model then generates output block-by-block in a streaming fashion, estimating the posterior using the causality-based C.I. as:
\begin{multline}
P(Y^{\text{sds}}|X, X^{\text{spk}})= \prod_{b}~P(Y^{\text{sds}}_{b+1}|Y^{\text{sds}}_{1:b}, \\
X_{1:b},\cancel{X_{b+1:B}},X^{\text{spk}}).
\label{eq:intro_sds_time_multiplex}
\end{multline}

Building on earlier turn-by-turn Chain-of-Thought (CoT) formulations for E2E SDS, recent work, namely SCoT, has investigated how structured intermediate reasoning can be integrated into blockwise duplex systems~\cite{SCoT}. The key idea is to introduce intermediate representations, such as ASR transcripts and textual responses, that are temporally aligned with the speech signal.
Let the corresponding aligned transcript and system response for the $b^{\text{th}}$ block (See S.~\ref{subsec:duplex_SDS}) be $A^{\text{asr}}_b$ and $A^{\text{res}}_b$ respectively.
Incorporating these alignments, the blockwise dialogue policy (Eq.~\ref{eq:intro_sds_time_multiplex}) can be augmented to condition on partial reasoning states (similar to (Eqs.~\eqref{eq:turn_cot_final_formulation}-\eqref{eq:turn_cot_t2t_formulation}) as shown :
\begin{align}
P(Y^{\text{sds}}|X, X^{\text{spk}}) &\approx \prod_{b} P(Y^{\text{sds}}_{b+1}|Y^{\text{sds}}_{1:b},X_{1:b}, \nonumber \\
&\quad \hat{A}^{\text{asr}}_{1:b}, \hat{A}^{\text{res}}_{1:b+1},X^{\text{spk}}).\label{12_duplex_sds_eq:cot}
\end{align}
The intermediate text response is predicted as:
\begin{multline}
    \hat{A}^{\text{res}}_{b+1} = \argmax P({A}^{\text{res}}_{b+1}|Y^{\text{sds}}_{1:b},X_{1:b},\\
    \hat{A}^{\text{asr}}_{1:b}, \hat{A}^{\text{res}}_{1:b})\label{12_duplex_sds_eq:t2t}
\end{multline}
and the aligned ASR transcript for each block is:
\begin{multline}
\hat{A}^{\text{asr}}_{b} = \argmax P({A}^{\text{asr}}_{b}|Y^{\text{sds}}_{1:b},X_{1:b},\\\hat{A}^{\text{asr}}_{1:b-1}, \hat{A}^{\text{res}}_{1:b}).\label{12_duplex_sds_eq:asr}
\end{multline}
At each block, SCoT performs a structured three-stage decoding process, similar to S.~\ref{subsec:turn-by-turn}.
\subsection{DPO Data Preparation.}
\label{subsec:appendix_dpo_data_prep}

\paragraph{Semantic Reward:} For each dialogue turn, the CoT model produces $10$ candidate text responses using top-$k$ sampling ($k=10$), ensuring sufficient lexical and semantic diversity. We analyze the empirical distribution of the LLM judge scores across all candidates using histograms (Figures~\ref{fig:llm-score-distribution} and ~\ref{fig:autobleu-distribution} in Appendix), which reveal natural separation between coherent and incoherent responses. Based on this distribution, we define positive samples as those with an LLM score greater than $\tau_{\text{pos}}=6$ and AutoBLEU less than $\delta_{\text{low}}=30$, indicating semantically meaningful and non-repetitive outputs. Conversely, negative samples are those with an LLM score below $\tau_{\text{neg}}=5$ or AutoBLEU exceeding $\delta_{\text{high}}=30$, capturing incoherent, off-topic, or repetitive candidates.

\paragraph{Acoustic and Emotion Reward:}
For each candidate text response described above, we generate multiple spoken realizations using top-$k$ sampling ($k=10$), resulting in a set of candidate speech responses per turn. To construct intelligibility-based preference pairs, we retain samples satisfying an upper intelligibility constraint $\tau_{\text{wer}} = 0.25$, and select negative examples whose WER exceeds that of the positive sample by a margin $\delta_{\text{wer}} = 0.05$. In parallel, we construct audio-quality preference pairs using UTMOS scores. 
For emotion reward pairs, we use {Emo2Vec}~\cite{ma2023emotion2vec} to extract emotion representations from both the generated speech outputs and the corresponding ground-truth human speech. For preference construction, the candidate with the maximum emotion similarity is selected as the positive sample, while the candidate with the minimum similarity is selected as the negative sample, subject to a minimum margin of $\delta_{\text{emo}}=2\%$ between the two scores.

\begin{figure}[t]
\centering
    \includegraphics[width=\linewidth]{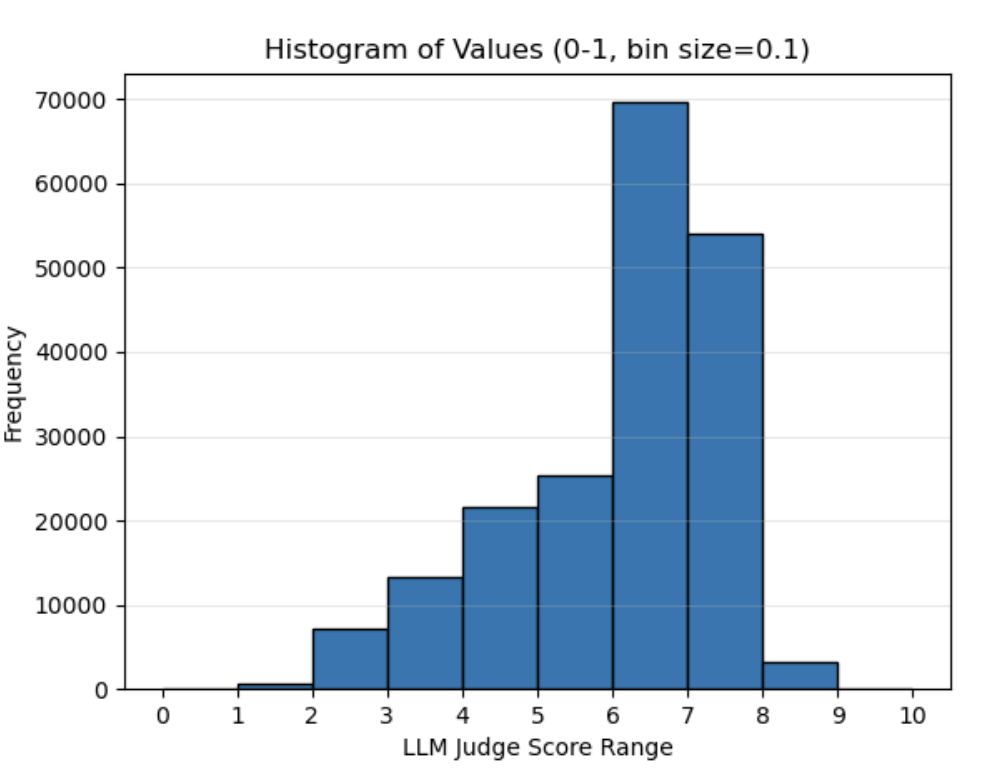} %
\caption{Distribution of LLM judge score over candidate responses}
\label{fig:llm-score-distribution}
\end{figure}
\begin{figure}[t]
\centering
    \includegraphics[width=\linewidth]{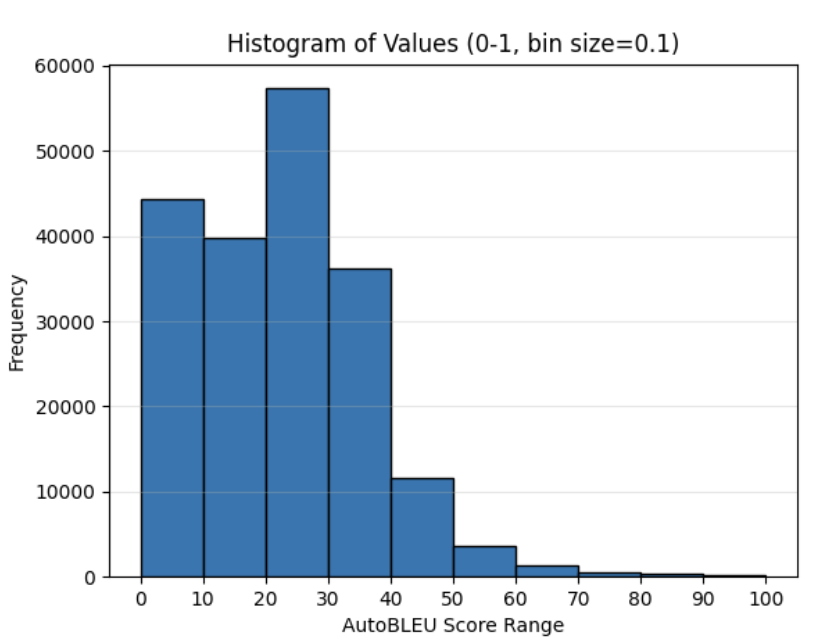} %
\caption{Distribution of AutoBLEU score over candidate responses}
\label{fig:autobleu-distribution}
\end{figure}

\begin{table*}[t]
\centering
\caption{Qualitative examples where RLAIF post-training improves response quality. Compared to the baseline, RLAIF-generated responses better address the user’s question, reduce topic drift, and avoid repetitive or incoherent continuations.}
\small
\setlength{\tabcolsep}{4pt}
\begin{tabular}{p{2.9cm} p{4cm} p{4cm} p{4cm}}
\toprule
\textbf{Dialogue Context} & \textbf{Reference Answer} & \textbf{w/o RLHF} & \textbf{RLHF} \\
\midrule

\textbf{Example 1} \newline
User: Do you have class on Friday?
&
Well, I have an exam eight thirty to ten thirty. Mike is taking the whole day off. Mike wanted to leave at like nine o'clock in the morning, but...
&
No, we're going to have lunch in the afternoon, but we had to do something with our hair because we got a lot of hair.
&
No, we don't have class on Friday. We have class on Saturday, so I'm going to go to the wedding.
\\
\midrule

\textbf{Example 2} \newline
User: We are starting the transition. I told the students what the new plan was...
&
What do you mean, oh, about you are leaving?
&
Yeah, I know what you mean.
&
Yeah, I know, I know. I'm just trying to figure out what to do.
\\
\midrule

\textbf{Example 3} \newline
User: I have been there for two weeks already.
&
Oh, my gosh! So, who?
&
Oh, okay. I mean, you start it on the fourth day if you want.
&
Oh, well, I'm glad you're doing well.
\\

\bottomrule
\end{tabular}
\label{tab:qualitative_better}
\end{table*}
\begin{table*}[t]
\centering
\caption{Qualitative examples where RLAIF post-training produces overly generic or safe responses that fail to advance the conversation, despite being coherent.}
\small
\setlength{\tabcolsep}{4pt}
\begin{tabular}{p{2.9cm} p{4cm} p{4cm} p{4cm}}
\toprule
\textbf{Dialogue Context} & \textbf{Reference Answer} & \textbf{w/o RLHF} & \textbf{RLHF} \\
\midrule

\textbf{Example 1} \newline
User: The problem I had when I went into teaching...
&
What grade were you teaching?
&
Yeah, I think that's good, too. I think that that should be, that's the ideal teacher...
&
Yeah, I think that's true.
\\
\midrule

\textbf{Example 2} \newline
User: That is what my sister is like. I want a meatball sandwich from Wawa. Let us walk up. I am like.
&
Is your sister doing that thing you know, hosing down the backyard?
&
I think I would have to go with my sister. I don't go to work with my brother...
&
Yeah, I know, I know.
\\

\bottomrule
\end{tabular}
\label{tab:qualitative_worse}
\end{table*}
\subsection{Experiment Setup}
\label{subsec:appendix_experiment_setup}
Our models are implemented in PyTorch, with all experiments conducted using the ESPnet toolkit~\cite{espnet,ESPnet-SLU,arora2025espnet}.
We use the pre-trained SpeechLM from~\cite{jinchuan2024speechlm}, which is initialized from the SmolLM2 1.7B text LLM.
We adopt the delay-interleaving architecture~\cite{musicgen} for multi-stream language modeling.

The SpeechLM is pre-trained for 500k updates using a warm-up–decay learning rate schedule, with each batch containing approximately 3,840 seconds of audio and 1.1M text tokens.
Training is performed on a large multimodal corpus comprising 213k hours of speech audio and 292B text tokens.
For audio tokenization, we use ESPnet-Codec~\cite{shi2024espnet}\footnote{\url{https://huggingface.co/ftshijt/espnet_codec_dac_large_v1.4_360epoch}}
 for codec tokenization and XEUS~\cite{chen2024towards}\footnote{\url{https://huggingface.co/espnet/xeus}
, a K-means tokenizer trained on the last-layer representations with 5k clusters} for SSL tokenization. Specifically, codec and SSL tokens are concatenated frame-by-frame.

For decoding, ASR outputs are generated using greedy search followed by hallucination detection and removal, as in~\cite{arora2025cotsds}.
Text response generation uses top-k sampling ($k=30$, temperature $=0.8$), with the same post-processing procedure~\cite{arora2025cotsds}.
For speech response generation in our CoT models, we again use top-k sampling (consistent with the text setting) and apply an additional intelligibility-based post-processing step following~\cite{arora2025cotsds}. Specifically, we generate ten candidate speech samples and transcribe each using Whisper~\cite{whisper}. We compute the WER of each transcription against the model’s own predicted text response, $\hat{S}^{\text{res}}_{k}$ (Eq.~\eqref{eq:turn_cot_t2t_formulation}) for turn-by-turn systems and $\hat{A}^{\text{res}}_{b+1}$ (Eq.~\ref{12_duplex_sds_eq:t2t}) for duplex systems, as the reference, and select the candidate with the lowest WER.

During inference in blockwise duplex systems, we additionally constrain the ASR transcript and text response to a maximum of 25 words and limit the speech output to the block duration (i.e., 2 seconds), following SCoT~\cite{SCoT}.
To ensure fair comparison between turn-by-turn and duplex systems, we aggregate speech outputs across blocks within each turn and compute quality metrics on the combined output.
Finally, for the baseline Moshi system, we provide each turn-level utterance as input and append 20 seconds of silence to ensure response generation is triggered.

All models are trained using 4 NVIDIA H200 GPUs. Further, we split DPO data shown in Table~\ref{tab:dpo_data_stats}, into 99:1 ratio for training / validation while training our preference learning models. Model hyperparameters are shown in Table~\ref{tab:opuslm_posttrain_params}. All results are reported from a single training run.
We will publicly release data processing, training and inference details.
\begin{table*}[h!]
\centering
\caption{RLHF Post-training Parameters. Hyperparameters found based on performance on validation set.}
\begin{tabular}{|p{5cm}|p{7cm}|}
\hline
\textbf{Parameter} & \textbf{Value} \\
\hline
gradient\_accumulation\_steps & 1 \\
\hline
epochs & 2 \\
\hline
gradient\_clipping & 100.0 \\
\hline
bf16 enabled & true \\
\hline
optimizer type & Adam \\
\hline
optimizer lr & 0.0000006 \\
\hline
optimizer betas & [0.9, 0.95] \\
\hline
optimizer eps & 1e-8 \\
\hline
optimizer weight\_decay & 3e-7 \\
\hline
optimizer adam\_w\_mode & true \\
\hline
scheduler type &"WarmupCosineLR" \\
\hline
scheduler warmup\_type & linear \\
\hline
scheduler total\_num\_steps & 9000 \\
\hline
scheduler warmup\_num\_steps & 100 \\
\hline
scheduler warmup\_min\_lr & 0 \\
\hline
scheduler warmup\_max\_lr & 0.00001 \\
\hline
\end{tabular}
\label{tab:opuslm_posttrain_params}
\end{table*}
\clearpage
\clearpage
\subsection{LLM Prompt}
\label{box:llm_judge_prompt}
\begin{tcolorbox}[
    width=\textwidth,
    colback=gray!5,
    colframe=black,
    title={Prompt Used for LLM-Based Evaluation and DPO Data Construction},
]
\footnotesize
\begin{verbatim}
Please rate the response from the voice dialogue system based on the human reference response
and input user utterance and following criteria (1-10 points), and provide a brief evaluation:
1. Relevance: Is the response relevant to the query? Is the content related?
2. Accuracy: Does the response correctly address the user's query and provide accurate information?
3. Completeness: Does the response comprehensively cover all aspects of the query?
4. Conversational Nature: Is the response easy to understand, concise, clear, and fluent?
Output in JSON format:
{
"Strengths": "Positive aspects of the response",
"Weaknesses": "Negative aspects of the response",
"Overall Evaluation": "Overall assessment of the response",
"Total Score (out of 10, directly provide the score)": ""
}
<Dialogue Context>
Reference: <Reference Response>
Agent: <SDS Response>
\end{verbatim}
\end{tcolorbox}
\end{document}